\documentclass{svproc}
\usepackage{amsfonts,amsmath,
algorithm,amssymb,graphicx,mathrsfs}
\RequirePackage[colorlinks,citecolor=blue]{hyperref}
\allowdisplaybreaks

\usepackage{url}

\spnewtheorem{assumption}{Assumption}{\bfseries}{\itshape}

\begin{document}
\mainmatter              
\title{On regret bounds for continual single-index learning}
\titlerunning{Regret bounds for continual single-index learning}  
%
\author{T. Tien Mai }
\authorrunning{T.T. Mai} 
%
\tocauthor{T. Tien Mai}
\institute{Department of Mathematical Sciences,
	\\
	Norwegian University of Science and Technology,
	Trondheim, Norway.
	\\ \email{the.t.mai@ntnu.no}}

\maketitle              

\begin{abstract}
In this paper, we generalize the problem of single-index model to the context of continual learning in which a learner is challenged with a sequence of  tasks one by one and the dataset of each task is revealed in an online fashion.  We propose a randomized strategy that is able to learn a common single-index (meta-parameter) for all tasks and a specific link function for each task. The common single-index allows to transfer the information gained from the previous tasks to a new one.  We provide a rigorous theoretical analysis of our proposed strategy by proving some regret bounds under different assumption on the loss function.  
\end{abstract}

\section{Introduction}

Recently,  studying of learning algorithms in the setting in which the tasks are presented sequentially has received a lot of attention, see e.g. \cite{pentina2014pac,chen2018lifelong,alquier2017regret,finn2019online,nguyen2018variational,denevi2018learning,wu2019lifelong}
among others. This setting is often referred to as contunual learning,  also called as learning-to-learn or incremental learning \cite{thrun,Baxter2000,alquier2017regret}. Clearly, using information gained from previously learned tasks 
is useful and important for learning a new similar task. This is motivated from that human are able to learn a new task quite accurately by utilizing knowledge from previous learned tasks.

In order to reuse the information from previous tasks, the new task must share some commonalities with previous ones.  In this work, we consider that different tasks share a common feature representation space. This direction has been explored by various works, e.g. \cite{ruvolo2013ella,PenBenDavid15,alquier2017regret,wu2019lifelong,denevi2018incremental,denevi2019online} and is natural for classification and regression problem. More precisely, different predictor for each task is built on top of a common representation in order to make predictions.

In this paper,  we extend the single-index model \cite{mccullagh1989generalized} to the learning-to-learn setting. More specifically, we assume that the tasks share a common single-index (meta-parameter) in this problem.  The predictor is constructed on top of this common single-index through a task-specific link functions (predictors).  This grants the learner to reuse/transfer the knowlegde (the commonality) learned from previous tasks to a new task through the common single-index. Moreover, the learner still has the ability to deal with the differency between tasks through a task-specific link function.

Continual learning can be casted as a generalization of online learning and a standard way to provide theoretical guarantees for online algorithms is a regret bound.  This bound measures the discrepancy between the prediction error of the forecaster and the error of an ideal predictor. We extend the EWA-LL meta-procedure in \cite{alquier2017regret} to our continual single-index learning problem.  Through this procedure, we provide the regret bounds for continual learning single-index. These theoretical analysis show that it is possible to learn such model in a continual context.

Interestingly, as a by-product from our work that is to provide an example of a within-task algorithm, we develop an online algorithm for learning single-index model in an online setting.  More specifically, it is based on the exponentially weighted aggregation (EWA) procedure for online learning, see e.g. \cite{cesa2006prediction} and references therein. We also provide a regret bound for this algorithm which is also novel in the context of online single-index learning.

The paper is structured as follow.  In Section \ref{notations} we introduce the continual learning context and then extend the single-index model to this context. After that, we present a meta algorithm for learning the continual single-index model based on EWA-LL procedure. The regret bound analysis is given in Section \ref{sc_regret}.  A within-task online algorithm for single-index model and its regret bound is presented in Section \ref{sc_withintask}. Some discussion and conclusion are given in Section \ref{sc_conclusion}. All technical proofs are given in Section \ref{proofs}.

\section{Continual single-index setting}
\label{notations}
\subsection{Setting}
We assume that, at each time step $t\in\{1,\dots,T\}$, 
the learner is challenged with a task sequentially, corresponding to a dataset
$$ \mathcal{S}_t = 
\{ (x_{t,1},y_{t,1}),\dots,(x_{t,n_t},y_{t,n_t}) \}
\in 
(\mathcal{X} \times \mathcal{Y})^{n_t}, 
n_t \in {\mathbb N}.
$$
Furthermore, we assume that the dataset $\mathcal{S}_t$ is itself revealed sequentially, that is, at each inner step $i \in \{1,\dots,n_t\}$:
\begin{itemize}
 \item the object $x_{t,i}$ is revealed and the learner has to predict $y_{t,i}$ by $\hat{y}_{t,i} $;
 \item then $y_{t,i}$ is revealed and the learner incurs the loss $\hat{\ell}_{t,i} := \ell(\hat{y}_{t,i},y_{t,i})$.
\end{itemize}

Let $f:\mathcal{X}\rightarrow \mathcal{Y}$ be a predictor,  where $\mathcal{Y} = \mathbb{R}$ for regression and $\mathcal{Y}=\{-1,1\}$ for binary classification.  Put $\hat{y}_{t,i}:= f(x_{t,i}) $ denote the prediction for $y_{t,i}$.  

As we want to transfer the information (a common data representation) gained from the previous tasks to a new one. Formally, we let $\mathcal{Z}$ be a set and prescribe a set $\mathcal{G}$ of feature maps (also called {\em representations}) $g:\mathcal{X}\rightarrow \mathcal{Z}$, and a set $\mathcal{H}$ of functions $h:\mathcal{Z}\rightarrow \mathbb{R}$. We shall design an algorithm that is useful when there is a function $g\in\mathcal{G}$, common to all the tasks, and task-specific functions $h_1,\dots,h_T$ such that 
$
f_t= h_t \circ g
$
is a good predictor for task $t$, in the sense that the corresponding prediction error (see below) is small.

\begin{figure}
\centering
\includegraphics[scale=.3]{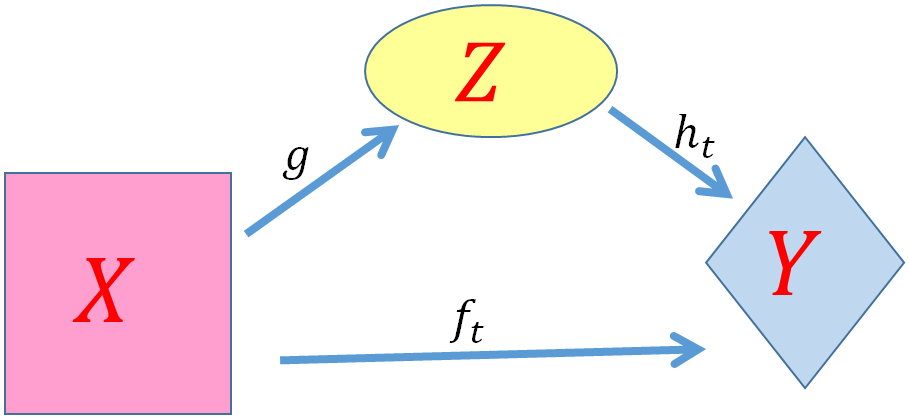}
\caption{The predictor $f_t$ is built on top of a representation $g$ and a task-specific function $h_t$.}
\end{figure}

In the single index model, let the set $\mathcal{X}=\mathcal{Z}=\mathbb{R}^d$, and we define $\mathcal{G}= \{ x\mapsto \theta^\top x, \theta \in \mathbb{R}^d \}$ linear functions on ${\cal X}$.  Furthermore, let
${\cal H}$ be a set of $L_2$-Lipschitz univariate measurable functions on $\mathbb{R}$.  Recall that our predictor here is of the form 
$$
f_t (x_{t,i} ) = h_t (\theta^\top x_{t,i} ) .
$$ 
The goal is to learn the common single-index vector (meta-parameter) $\theta$ for all tasks and the link function (task's specific predictor) $h_t$ for each task $t$.

The predictor can be interpreted as that the predictor changes only in the direction $\theta $ (single-index), and the way it changes in this direction is defined by the link function $h_t$.

\begin{remark}
The single-index model \cite{mccullagh1989generalized} is known as a particularly useful variation of the linear model. This model has been applied to a variety of fields, such as discrete choice analysis in econometrics and dose-response models in biometrics, where high-dimensional regression models are often employed.  See for example \cite{horowitz2012semiparametric,mcaleer2008single,ichimura1993semiparametric,lopez2009single}.
\end{remark}

Noted that the task $t$ ends at time $n_t$ and the average prediction error at that point is $ \frac{1}{n_t} \sum_{i=1}^{n_t}\hat{\ell}_{t,i}$. This process is repeated for each task $t$, so that at the end of all the tasks, the average error is
$
\frac{1}{T} \sum_{t=1}^T \frac{1}{n_t} \sum_{i=1}^{n_t}\hat{\ell}_{t,i}.
$
Our principal objective is to design a procedure (meta-algorithm) that is able to learn the common single-index vector (meta-parameter) $\theta$ for all tasks and the task's specific predictor $h_t$ for each task $t$ and control the {\em (compound) regret} of our procedure
\begin{align*}
\mathcal{R}
 :=
\frac{1}{T}\sum_{t=1}^T \frac{1}{n_t} \sum_{i=1}^{n_t}\hat{\ell}_{t,i}
-
\inf_{\theta}
\frac{1}{T}\sum_{t=1}^T
\inf_{h_t \in\mathcal{H}} \frac{1}{n_t} \sum_{i=1}^{n_t}
   \ell\big(  h_t(\theta^\top x_{t,i} ) ,   y_{t,i}\big).
\end{align*}

\subsection{A randomized strategy for continual single-index learning}

The EWA-LL meta-algorithm proposed in \cite{alquier2017regret} based on the exponentially weighted aggregation (EWA) is a general procedure in lifelong learning. Here, we propose an application of this algorithm to the context of single-index learning. The details of our proposal algorithm is outlined in Algorithm \ref{algo_ewa}.

\begin{algorithm}[!ht]
\caption{EWA-LL for continual single-index learning}
\begin{description}
\item[Data] A sequence of datasets $ \mathcal{S}_t = \big((x_{t,1},y_{t,1}),\dots,(x_{t,n_t},y_{t,n_t})\big) ,  1\leq t\leq T$; the points within each dataset are also given sequentially.
\item[Input] A prior $\pi_1$, a learning parameter $\eta>0$ and a learning algorithm for each task $t$ which, for any single-index $\theta $ returns a
sequence of predictions $\hat{y}_{t,i}^\theta $ and suffers a loss
$$ \hat{L}_t(\theta ) := \frac{1}{n_t} \sum_{i=1}^{n_t}
       \ell\left(\hat{y}_{t,i}^\theta ,y_{t,i}\right). $$
\item[Loop] For $t=1,\dots,T$
\begin{description}
\item[i] Draw $\hat{\theta }_t \sim \pi_t$.
\item[ii] Run the within-task learning algorithm on $ \mathcal{S}_t $ and suffer
loss $ \hat{L}_t(\hat{\theta }_t)$.
\item[iii] Update
 $$
 \pi_{t+1}({\rm d}\theta ) := \frac{\exp(-\eta \hat{L}_t(\theta ))
                  \pi_{t}({\rm d}\theta ) }{\int \exp(-\eta \hat{L}_t(\gamma))
                  \pi_{t}({\rm d}\gamma) }.
 $$
\end{description}
\end{description}
\label{algo_ewa}
\end{algorithm}

More specifically, the algorithm  \ref{algo_ewa} is based on the
exponentially weighted aggregation (EWA) procedure, see e.g. \cite{cesa2006prediction,audibert2006randomized} and references therein. It updates a probability distribution $\pi_t$ on the set of single-index representation ${\cal G} $ before the encounter of task $t$. It is noticed that this procedure allows the user to freely choose the within-task algorithm (step {\bf ii}) to learn the task-specific link function $h_t$,   which does not even need to be the same for each task.

Furthermore,  the step {\bf i} is crucial during the learning procedure, because to draw $\hat{\theta }_t $ from $\pi_t$ is not straightforward and varies in different specific situation.
While the effect of Step {\bf iii} is that any single-index $\theta  $ which does not perform well on task $t$, is less likely to be reused on the next task.

\section{Regret bounds}
\label{sc_regret}

\subsection{Bound with expectation}
We make the following assumptions on our model.
\begin{assumption}
\label{assum_1}
We assume that $\|\theta\|_1 = 1 $ and $  \| x_{t,i}\|_2 \leq M <+\infty $.
\end{assumption}

\begin{assumption}
\label{assum_2}
We assume that the loss $\ell $ is $L_1$-Lipschitz with respect to its first component, i.e,  there exists $L_1 >0$ such that
$$
| \ell (a_1, \cdot ) - \ell (a_2,  \cdot )  | \leq  L_1 |a_1 - a_2|.
$$
We further assume that $\ell(x,\cdot) \in [0,C],\forall x $.
\end{assumption}

Assume that we have some within-task algorithms that learn $h_t$ at each time $t$. And 
$$
\beta(n_t):= \sup_{\|\theta\|_1 = 1 }
\left\lbrace 
 \frac{1}{n_t} \sum_{i=1}^{n_t}\hat{\ell}_{t,i}
-
\inf_{h_t \in\mathcal{H}} \frac{1}{n_t} \sum_{i=1}^{n_t}
   \ell\big(  h_t(\theta^\top x_{t,i} ) ,   y_{t,i} \big)
   \right\rbrace
    <+\infty ,
$$
$\beta(n_t)$ being an upper bound of the within-task-regret of a within-task algorithm that learns $h_t$. We will detail one possible such algorithm in Section \ref{sc_withintask}.

Let $\pi_1$ be uniform on the unit $\ell_1 $-ball.  We note that as Algortihm \ref{algo_ewa} is a randomized algorithm, we first provide a bound on the expected regret. A simple result for continual single-index learning is given in the following theorem.
\begin{theorem}
\label{thm_main}
Under the Assumptions \ref{assum_1} and \ref{assum_2}, we have
\begin{multline*}
 \frac{1}{T} \sum_{t=1}^T \mathbb{E}_{\hat{\theta} \sim \pi_t} 
\left[  \frac{1}{n_t} \sum_{i=1}^{n_t}  \hat{\ell}_{t,i} \right]
  -
  \inf_{\|\theta\|_1 = 1}  \frac{1}{T} \sum_{t=1}^T
  \inf_{h_t \in\mathcal{H}} \frac{1}{n_t}\sum_{i=1}^{n_t}
 \ell(h_t(\theta^\top x_{t,i}),y_{t,i})
 \\
 \leq
  \frac{c_{(L_1, L_2, C,M)} d\log(T)+ 2d\log\left(d\right)  }{\sqrt{T}} 
+   \frac{1}{T}\sum_{t=1}^T \beta(n_t) .
\end{multline*}
where $ c_{(L_1, L_2, C,M)} $ is a universal constant that depends only on $ L_1 , L_2, M $ and $C$.
\end{theorem}
The proof relies on an application of Theorem 3.1 in \cite{alquier2017regret}.
We postpone the proof to Section~\ref{proofs}.

\subsection{Uniform bound}
Now, under additional assumption that the loss function is convex with respect to (w.r.t.) its first component,   it is possible to obtain a uniform regret bound.  However, rather than using a random draw that $ \hat{\theta }_t \sim \pi_t $ as in Step \textbf{i} of Algorithm \ref{algo_ewa}, we need to consider an aggregation step for predicting that is 
\begin{equation}
\label{integral}
\hat{y}_{t,i}
 = \int h_t (\theta^\top x_{t,i} )   \pi_t({\rm d} \theta).
\end{equation}
The uniform regret bound is presented in the following theorem.
\begin{theorem}
\label{thm_uniform}
 Under the assumptions of Theorem \ref{thm_main} and the loss function is convex w.r.t its first argument, we have
\begin{multline*}
 \frac{1}{T} \sum_{t=1}^T\frac{1}{n_t} \sum_{i=1}^{n_t} 
 \ell(  \hat{y}_{t,i} ,y_{t,i})
  -
  \inf_{\|\theta\|_1 = 1}  \frac{1}{T} \sum_{t=1}^T
  \inf_{h_t \in\mathcal{H}} \frac{1}{n_t}\sum_{i=1}^{n_t}
 \ell(h_t(\theta^\top x_{t,i}),y_{t,i})
 \\
 \leq
  \frac{c_{(L_1, L_2, C,M)} d\log(T)+ 2d\log\left(d\right)  }{\sqrt{T}} 
+   \frac{1}{T}\sum_{t=1}^T \beta(n_t) .
\end{multline*}
where $ c_{(L_1, L_2, C,M)} $ is a universal constant that depends only on $ L_1 , L_2, M $ and $C$.
\end{theorem}
\begin{proof}
We have that
$$
  \frac{1}{n_t} \sum_{i=1}^{n_t} 
  \ell (\hat{y}_{t,i}, y_{t,i})  
  =  
  \frac{1}{n_t} \sum_{i=1}^{n_t} 
  \ell \left( \int h_t(\theta^\top x_{t,i} ) \pi_t(d\theta ) , y_{t,i} \right)  .
$$
As the loss is convex w.r.t its first component,  Jensen's inequality leads to
$$
  \frac{1}{n_t} \sum_{i=1}^{n_t} 
  \ell \left( \int h_t(\theta^\top x_{t,i} ) \pi_t(d\theta ) , y_{t,i} \right) 
  \leq
   \int   \frac{1}{n_t} \sum_{i=1}^{n_t} 
  \ell \left(h_t(\theta^\top x_{t,i} ), y_{t,i} \right)  \pi_t(d\theta ). 
$$
The proof completes by applying Theorem \ref{thm_main}.
\end{proof}

\begin{remark}
Our regret bound is typically of order $\log(T)/\sqrt{T} $ , which tends to $0 $ as the number of tasks, $T$ increases.
\end{remark}

\begin{remark}
Noted that if all the tasks have the same sample size, that is $n_t = n $ for all $t $, then $  \frac{1}{T}\sum_{t=1}^T \beta(n_t) = \beta(n) $ and thus the analysis will not be changed.  Here after,  to ease our presentation, we assume that all the tasks have the same sample size, that is $n_t = n , \forall t$.
\end{remark}

\subsection{Bounds with Monte Carlo approximation}
\label{sc_bound_MCMC}
In practice, for an infinite set $\mathcal{G}$ we are not able to run simultaneously the within-task algorithm for all single-index $\theta $. So, we cannot compute the prediction~\eqref{integral} exactly. A possible
strategy is to draw $N$ elements i.i.d. from $\pi_t$,
say $\hat{\theta}_{t}^{(1)},\dots,\hat{\theta}_{t}^{(N)}$, and to replace~\eqref{integral} by its Monte Carlo approximation
$$ \hat{y}_{t,i}^{(N)}
 = \frac{1}{N} \sum_{j=1}^N  h_t ( \hat{\theta}_{t}^{(j)\top} x_{t,i} ) . $$
Let's call MC-EWA this new version.

\begin{algorithm}[!ht]
\caption{MC-EWA for continual single-index learning with convex loss}
\begin{description}
\item[Data and Input] as in Algorithm \ref{algo_ewa}.
\item[Loop] For $t=1,\dots,T$
\begin{description}
\item[i] Draw $\hat{\theta}_{t}^{(1)},\dots,\hat{\theta}_{t}^{(N)}$ i.i.d from $\pi_t$.
\item[ii] Run the within-task learning algorithm $ \mathcal{S}_t $ for each $ \hat{\theta}_{t}^{(j)} $
and return as predictions:
$$ \hat{y}_{t,i}^{(N)}
 = \frac{1}{N} \sum_{j=1}^N  h_t ( \hat{\theta}_{t}^{(j)  \top} x_{t,i} ) . $$
\item[iii] Update
 $
 \pi_{t+1}({\rm d} \theta) := \frac{\exp(-\eta \hat{L}_t(\theta))
                  \pi_{t}({\rm d} \theta) }{\int \exp(-\eta \hat{L}_t(\gamma))
                  \pi_{t}({\rm d}\gamma) }.
 $
\end{description}
\end{description}
\end{algorithm}

In order to analyze the performance of this algorithm, we can directly use Theorem~\ref{thm_uniform}. We only have to control the
discrepancy between the theoretical integral with respect to $\pi_t$ and
the corresponding empirical mean.
Hoeffding's inequality leads to
$$
\frac{1}{N} \sum_{j=1}^N \hat{L}_t(  \hat{\theta}_{t}^{(j)}  )
\leq 
\mathbb{E}_{ \theta \sim \pi_t} [\hat{L}_t (\theta)] 
+ 
C \sqrt{\frac{\log\left(
      \frac{1}{\delta}\right)}{2N}}
$$
with probability at least $1-\delta$. A union bound over the $T$ tasks leads
to the following result directly.
\begin{corollary}
\label{cor:online:w:online}
 Assuming that the assumptions of Theorem \ref{thm_uniform} are hold. Then, with probability at least $1-\delta$ over the drawing of all the $   \hat{\theta}_{t}^{(j)}  $'s,
 \begin{multline*}
 \frac{1}{T} \sum_{t=1}^T \frac{1}{n_t} \sum_{i=1}^{n_t}
  \ell \left(  \hat{y}_{t,i}^{(N)},y_{t,i}  \right)
  -
  \inf_{\|\theta\|_1 = 1}  \frac{1}{T} \sum_{t=1}^T
  \inf_{h_t \in\mathcal{H}} \frac{1}{n_t}\sum_{i=1}^{n_t}
 \ell(h_t(\theta^\top x_{t,i}),y_{t,i})
 \\
 \leq
  \frac{c_{(L_1, L_2, C,M)} d\log(T)+ 2d\log\left(d\right)  }{\sqrt{T}} 
+   \frac{1}{T}\sum_{t=1}^T \beta(n_t)  + C \sqrt{\frac{\log\left(
      \frac{T}{\delta}\right)}{2N}}.
 \end{multline*}
\end{corollary}

In the next Section, we provide an example of a within task online algorithm and derive its regret bound.

\section{A within-task algorithm} 
\label{sc_withintask}
\subsection{EWA for online single-index learning}
Here, we propose an online algorithm for learning within each task, detailed in Algorithm \ref{algo_withintask}. The algorithm is based on the EWA procedure on the space ${\cal H} \circ g$ for a prescribed single-index representation $g \in {\cal G}$, with $g(x) = \theta x $.  

\begin{algorithm}[H]
\caption{EWA for online single-index learning}
\begin{description}
\item[Data] A task $ \mathcal{S}_t = \big((x_{t,1},y_{t,1}),\dots,(x_{t,n_t},y_{t,n_t})\big) $ where the data points are given sequentially.
\item[Input] A learning rate $\zeta>0$; 
\\a prior distribution $\mu_1$ on $\mathcal{H}$.
\item[Loop] For $i=1,\dots,n_t$,
\begin{description}
\item[i] Predict $\hat{y}_{t,i}^\theta = \int_{\mathcal{H}}
 h  (\theta^\top x_{t,i}) \mu_i ({\rm d} h)$,
\item[ii] $y_{t,i}$ is revealed, update
$$
 \mu_{i+1}({\rm d}h) = \frac{\exp(-\zeta \ell( \hat{y}_{t,i}^\theta  ,y_{t,i}))
    \mu_{i}({\rm d}h)}{\int \exp(-\zeta \ell(u (\theta^\top x_{t,i}),y_{t,i}))
    \mu_{i}({\rm d}u)}. 
    $$
\end{description}
\end{description}
\label{algo_withintask}
\end{algorithm}

To learn $h_t$, we use Algorithm \ref{algo_withintask} and consider a structure for  $\mathcal{H} $. We consider,  for a positive interger $S $, the link function (task's specific predictor)
$$h_t \in \mathcal{H}_{S,C_2+1} :=\{ h\in \mathcal{H}: h= \sum_{j=1}^S\beta_j \phi_j, \sum_{j=1}^S j |\beta_j| \leq C_2 +1\} ,$$ 
where $ \{\phi_j\}_{j=1}^{\infty}$ is a dictionary of measurable 
functions, each $\phi_j $ is assumed to be defined on $ [-1,1] $ and to take values in $[-1,1]$.  The trigonometric system \cite{tsybakov2008introduction} is an example for this kind of dictionary, that is $ \phi_1(z) = 1, \phi_{2j} (z) = \cos(\pi j z) , \phi_{2j+1} (z) = \sin(\pi j z) $ with $j = 1,2,\ldots $ and $z \in  [-1,1] $.

Let
\begin{align*}
\mathcal{B}_S(C_2 + 1) := \{(\beta_1,\ldots,\beta_S)\in \mathbb{R}^S:  \sum_{j=1}^S j |\beta_j| \leq C_2 +1 \}.
\end{align*}
We define $\mu_1(dh) $ on the set $\mathcal{H}_{S,C_2+1} $ as the image of 
the uniform measure on $ \mathcal{B}_S(C_2+ 1) $ induced by the map $ (\beta_1,\ldots,\beta_S)\mapsto \sum_{j=1}^S\beta_j \phi_j $.

\begin{remark}
The choice of $ C_2 + 1 $ instead of $ C_2  $ in the definition of the prior support is just convenient for technical proofs. This  ensures that when the target $h_t$ belongs to $ \mathcal{H}_{S,C_2}$, then a small ball around it is contained in $ \mathcal{H}_{S,C_2+1}$. The integer $S$ should be understood as a measure of the ``dimension" of the link function $h_t$; the larger $S$, the more complex the function.
\end{remark}

Now, we are ready to provide a regret bound for Algorithm \ref{algo_withintask}. Remind that we assume that $n_t = n , \forall t$.
\begin{proposition}
\label{cor_single_index}
By choosing  $\zeta = \sqrt{\frac{ 8S}{C^2 n}}$ , we have
\begin{align*}
 \frac{1}{n} \sum_{i=1}^{n} \hat{\ell}_{t,i}
 -  
  \inf_{h_t \in   \mathcal{H}_{S,C_2+1} }
 \frac{1}{n}\sum_{i=1}^{n}
 \ell(h_t(\theta^\top x_{t,i}),y_{t,i})
 \leq
 a_{(L_1, S , C,C_2)}  \frac{ \log(n) }{\sqrt{n}}   ,
\end{align*}
where $ a_{(L_1, S , C,C_2)} $ is a universal constant that depends only on $ L_1 ,  S , C,  C_2$.
\end{proposition}

As the proof of the Proposition \ref{cor_single_index} is not straightforward, we postpone the proof to Section~\ref{proofs}.

In practice, for an infinite set $ \mathcal{H}_{S,C_2+1} $ we may not compute the prediction integral in Algorithm \ref{algo_withintask} exactly. Following Section \ref{sc_bound_MCMC}, a possible approximate method is to draw $N_1 $ elements i.i.d. from $ \mu_i $,
say $\hat{h}^{(1)}  ,\dots,\hat{h}^{(N_1)}$, and to replace the integral by its Monte Carlo approximation
$$
\hat{y}_{t,i}^\theta 
=
 \int_{\mathcal{H}}
 h  (\theta^\top x_{t,i}) \mu_i ({\rm d} h)
  \approx  
  \sum_{s=1}^{N_1}  \hat{h}^{(s)}  (\theta^\top x_{t,i}).
 $$
An analysis follows exactly the procedure as in Section \ref{sc_bound_MCMC} for this Monte Carlo approximation leads to an additional cost for Proposition \ref{cor_single_index} by $ C \sqrt{\\log\left( \frac{n}{\delta}\right)  / (2N_1)} $.

\subsection{A detailed regret bound}
We are ready to provide a full regret bound for continual single-index learning. The following result is obtained by plug in Proposition \ref{cor_single_index} into Theorem \ref{thm_uniform}.
\begin{corollary}
 Under the assumptions of Theorem \ref{thm_uniform} and Proposition \ref{cor_single_index},  we have
\begin{multline*}
 \frac{1}{T} \sum_{t=1}^T  \frac{1}{n} \sum_{i=1}^{n} 
  \hat{\ell}_{t,i} 
  -
  \inf_{\|\theta\|_1 = 1}\frac{1}{T}\sum_{t=1}^T
  \inf_{h_t \in\mathcal{H}_{S,C_2+1} }\frac{1}{n}\sum_{i=1}^{n}
 \ell(h_t(\theta^\top x_{t,i}),y_{t,i})
 \\
 \leq
 a_{(L_1, S , C,C_2)}  \frac{ \log(n) }{\sqrt{n}}  
 + 
 \frac{c_{(L_1, L_2, C)} d\log(T)+ 2d\log\left(d\right)}{\sqrt{T}} ,
\end{multline*}
where $ c_{(L_1, L_2, C,M)} $ is a universal constant that depends only on $ L_1 , L_2 , C,M$ and  $ a_{(L_1, S , C,C_2)} $ is a universal constant that depends only on $ L_1 ,  S , C,  C_2$.
\end{corollary}

Typically, we obtain a regret bound at the order of  $\log(n)/\sqrt{n} + \log(T)/\sqrt{T}  $.


\section{Discussion and Conclusion}
\label{sc_conclusion}

We presented a meta-algorithm for continual single-index learning and provided for the first time a fully online analysis of its regret.  We also provided an online algortihm for learning within task and proved its regret bound. This is novel to our knowledge.

A fundamental question remains open is to provide a more computationally efficient algorithm, such as approximations of EWA \cite{cherief2019generalization}, or fully gradient based algorithms as in~\cite{ruvolo2013ella,nguyen2018variational}.

\section{Proofs}
\label{proofs}
First, we state the following result that is useful for our proofs.

\begin{theorem}
\emph{ \cite[Theorem 3.1]{alquier2017regret}}
\label{thm:online:w:online}
If, for any $g \in {\cal G},  \ell(x)\in [0,C]$ and the within-task algorithm has a regret bound $ \beta(g,n_t)$, then
 \begin{multline*}
 \frac{1}{T} \sum_{t=1}^T \mathbb{E}_{{\hat g}_t \sim \pi_t}\left[
 \frac{1}{n_t}
 \sum_{i=1}^{n_t} \hat{\ell}_{t,i}
 \right]
 \leq
 \inf_{\rho} \Biggl\{
       \mathbb{E}_{g \sim \rho}\Biggl[
        \frac{1}{T}
       \sum_{t=1}^T
       \inf_{h_t\in\mathcal{H}}
       \frac{1}{n_t}
       \sum_{i=1}^{n_t}
        \ell\big(h_t\circ g(x_{t,i}),y_{t,i}\big)
      \\
      +  \frac{1}{T} \sum_{t=1}^T \beta(g,n_t)
       \Biggr]
     + \frac{\eta C^2}{8} + \frac{\mathcal{K}(\rho,\pi_1)}{\eta T}
 \Biggr\},
 \end{multline*}
 where the infimum is taken over all probability measures $\rho$ and $\mathcal{K}(\rho,\pi_1)$ is the Kullback-Leibler divergence between
$\rho$ and $\pi_1$.
\end{theorem}

\subsection{Proof of Theorem~\ref{thm_main}}
\begin{proof}
Let $\theta^*$ denote a minimizer of the optimization problem
$$
\min_{\|\theta\|_1 = 1}\frac{1}{T}
\sum_{t=1}^T\inf_{h_t \in\mathcal{H}}
\frac{1}{n_t}\sum_{i=1}^{n_t}
 \ell ( h_t(\theta^\top x_{t,i}),y_{t,i}).
$$
We apply Theorem 3.1 in \cite{alquier2017regret} and 
upper bound the infimum w.r.t any $\rho$ by 
an infimum with respect to $\rho_\epsilon $ in the following 
parametric family 
\begin{align*}
\rho_{\epsilon}({\rm d} \theta)
\propto \mathbf{1}\{\|\theta-\theta^*\|_2 \leq \epsilon \}\pi_1(d\theta).
\end{align*}
where $\epsilon $ is a positive parameter. Note that when $ \epsilon$ is small, $\rho_\epsilon$ highly concentrates around $\theta^*$, but we will show this is at a price of an increase in ${\cal K}(\rho_\epsilon,\pi_1)$. The proof then proceeds in optimizing with respect to $ \epsilon$.

More specifically,  Theorem 3.1 in \cite{alquier2017regret} becomes
\begin{multline*}
 \frac{1}{T} \sum_{t=1}^T \mathbb{E}_{{\hat g}_t \sim \pi_t}\left[
 \frac{1}{n_t} \sum_{i=1}^{n_t} \hat{\ell}_{t,i}
 \right]
 \\
 \leq
 \inf_{ \epsilon} \Bigg\{
       \mathbb{E}_{\theta \sim \rho_{ \epsilon}}\bigg[
        \frac{1}{T}       \sum_{t=1}^T
       \inf_{h_t\in\mathcal{H}}
       \frac{1}{n_t}       \sum_{i=1}^{n_t}
        \ell ( h_t(\theta^\top x_{t,i}),y_{t,i})
+  \beta(n_t)
       \bigg]
     + \frac{\eta C^2}{8} + \frac{\mathcal{K}(\rho_{ \epsilon},\pi_1)}{\eta T}
 \Bigg\}.
 \end{multline*}

Furthermore, using the notation
\begin{align*}
h_t^* &:= \arg \inf_{h_t \in\mathcal{H}}
\frac{1}{n_t}\sum_{i=1}^{n_t}
 \ell  ( h_t(\theta^\top x_{t,i}),y_{t,i}) ,
\end{align*}
we get
\begin{multline*}
  \inf_{h_t \in\mathcal{H}}
\frac{1}{n_t}\sum_{i=1}^{n_t}
 \ell  ( h_t(\theta^\top x_{t,i}),y_{t,i})
 -
\frac{1}{n_t}\sum_{i=1}^{n_t}
 \ell ( h^*_t(\theta^* {}^\top x_{t,i}),y_{t,i})
 \\
\leq  
 \frac{1}{n_t}\sum_{i=1}^{n_t}
 \ell  ( h^*_t(\theta^\top x_{t,i}),y_{t,i})
 -
\frac{1}{n_t}\sum_{i=1}^{n_t}
 \ell ( h^*_t(\theta^* {}^\top x_{t,i}),y_{t,i}).
\end{multline*}
Under the condition on the loss, we have
\begin{align*}
\Big| \ell ( h^*_t(\theta^\top x_{t,i}),y_{t,i})
 -
 \ell ( h^*_t(\theta^* {}^\top x_{t,i}),y_{t,i}) \Big| 
 &      \leq
L\,  \Big|  h^*_t(\theta^\top x_{t,i})
 -
h^*_t(\theta^* {}^\top x_{t,i}) \Big|
\\
& \leq
L_1 L_2  | (\theta -\theta^*)^\top x_{t,i}|
\\
& \leq
\epsilon L_1 L_2  \| x_{t,i}\|_2.
\end{align*}
We obtain an upper-bound
\begin{multline*}
 \mathbb{E}_{\theta \sim \rho_{ \epsilon}}
        \frac{1}{T}       \sum_{t=1}^T
       \inf_{h_t\in\mathcal{H}}
       \frac{1}{n_t}       \sum_{i=1}^{n_t}
        \ell( h_t(\theta^\top x_{t,i}),y_{t,i})
 \\
 \leq
\inf_{\|\theta\|_1 = 1} \Biggl\{\frac{1}{T}
\sum_{t=1}^T\inf_{h_t \in\mathcal{H}}
\frac{1}{n_t}\sum_{i=1}^{n_t}
 \ell( h_t(\theta^\top x_{t,i}),y_{t,i})
 +  \epsilon \,L_1 L_2         \frac{1}{T} \sum_{t=1}^T
       \frac{1}{n_t}\sum_{i=1}^{n_t}  \| x_{t,i}\|_2
 \Biggr\}.
\end{multline*}

Now,  dealing with the Kullback-Leibler, we have 
$$\mathcal{K}(\rho_{ \epsilon},\pi_1) 
= - \log \pi_1(\{ \|\theta-\theta^*\|_2 \leq  \epsilon\}),$$
and
\begin{align*}
 \pi_1(\{ \|\theta-\theta^*\|_2\leq  \epsilon\}) 
& \geq   
\frac{\pi^{(d-1)/2}}{\Gamma((d+1)/2)} \epsilon^{(d-1)} \Bigg/ \frac{2^d}{d!}
\\
&  \geq 
\frac{ \epsilon^{(d-1)}}{\sqrt{\pi(d-1)}}\left(\frac{2\pi e}{d-1} \right)^{(d-1)/2}\Bigg/ \frac{2^d}{d!}
\\
&  \geq 
 \epsilon^{d-1} 2^{d-2}\frac{d!}{(d-1)^{d/2}}.
\end{align*}
Note that the first inequality follows by observing that, since 
$\pi_1 $ is the uniform distribution on the unit $\ell_1 $ ball, 
the probability to be calculated is greater or equal to the ration 
between the volume of the $(d-1)-$ball radius $ \epsilon $ over the volume of the unit $\ell_1 $ ball. The second inequality is just using the Stirling formula.  This trick had been used in \cite{mai2017pseudo}.

Consequently, we obtain
 \begin{align*}
\mathcal{K}(\rho_{ \epsilon},\pi_1) 
 \leq  
 (d-1)\log(1/ \epsilon) + \log\left(\frac{2^{d-2}d!}{(d-1)^{d/2}}\right)  .
\end{align*}
Therefore, Theorem 3.1 in \cite{alquier2017regret}  leads to
\begin{multline*}
 \frac{1}{T} \sum_{t=1}^T \mathbb{E}_{\theta \sim
  \pi_t}\left[ \frac{1}{n_t} \sum_{i=1}^{n_t} 
  \hat{\ell}_{t,i} \right]
  -
  \inf_{\|\theta\|_1 = 1}\frac{1}{T}
\sum_{t=1}^T\inf_{h_t \in\mathcal{H}}
\frac{1}{n_t}\sum_{i=1}^{n_t}
 \ell(h_t(\theta^\top x_{t,i}),y_{t,i})
 \\
 \leq
 \inf_{\epsilon} \left\{    \epsilon \,L_1 L_2  M  
 +  \frac{ (d-1)\log(1/\epsilon) }{\eta T}  
  \right\} +  \frac{\log\left(\frac{2^{d-2}d!}{(d-1)^{d/2}}\right) }{2\eta T}
+ \beta(n_t) + \frac{\eta C^2}{8} .
\end{multline*}
The choices $\epsilon = \sqrt{\frac{1}{T}}$ and 
$\eta = \frac{2}{C}\sqrt{\frac{1}{T}} $ make the right-hand side becomes
\begin{align*}
 \frac{L_1 L_2 M}{\sqrt{T}}
 +  \frac{Cd\log(T)+ \log\left(\frac{2^{d-2}d!}{(d-1)^{d/2}}\right) +C  }{4\sqrt{T}} 
+ \beta(n_t).
\end{align*}
The proof is completed by using the Stirling's approximation that $ \log(d!) \sim d\log(d) $.
\end{proof}

\subsection{Proof of Proposition~\ref{cor_single_index}}

\begin{proof}
We follows the same steps as in the proof of Theorem 1
in~\cite{audibert2006randomized}. First, we have, with $ \ell(h _u) := \ell(h  (\theta^\top x_{t,u}),y_{t,u}) $ that
\begin{align}
\label{renorm}
 \mu_{i}({\rm d}h) = \frac{\exp(-\zeta  \sum_{u=1}^{i-1}  \ell(h_u)  )
    \mu_{1}({\rm d}h)     }{  
      \int \exp(-\zeta  \sum_{u=1}^{i-1}  \ell(f_u)  )
    \mu_{1}({\rm d}f)        }
    = 
    \frac{\exp(-\zeta  \sum_{u=1}^{i-1}  \ell(h_u)  )
    \mu_{1}({\rm d}h)       }{   W_i}.   
\end{align}
where we introduce the notation $W_i$ for the sake of shortness. We denote $ \ell(h) :=   \ell(h  (\theta x_{t,i}),y_{t,i})) $ and put 
$$
E_i = \int  \ell(h)  \mu_{i} ({\rm d}h) 
= \mathbb{E}_{h_i \sim  \mu_{i}} [ \ell(h_i) ] .
$$
Using Hoeffding's inequality on the bounded random
variable $  \ell(h) \in[0,C]$ we have, for any $i $, that
$$
 \mathbb{E}_{h \sim  \mu_{i} } \left[  \exp\left\{ 
   \zeta (E_i-  \ell(h)    \right\} \right] 
   \leq 
   \exp\left\{ \frac{ C^2 \zeta^2}{8} \right\}
$$
which can be rewritten as
\begin{equation}
 \label{hoeffding}
\exp\left\{ - \zeta \mathbb{E}_{h \sim  \mu_{i} }[ \ell(h) ] \right\} 
   \geq 
   \exp\left( -\frac{C^2 \zeta^2}{8} \right)
 \mathbb{E}_{ h \sim  \mu_{i} } \left\{ \exp\left[ -
   \zeta  \ell(h) 
 \right] \right\}.
\end{equation}
Next, we note that
\begin{align*}
 \exp\left\{ - \zeta \sum_{i=1}^n \mathbb{E}_{h \sim  \mu_{i} }[ \ell(h) ]
 \right\}
  & = 
  \prod_{i=1}^n  \exp\left\{ - \zeta \mathbb{E}_{h \sim  \mu_{i} }[ \ell(h) ]
 \right\} 
 \\
  & \geq  
  \exp\left(- \frac{n C^2 \zeta^2}{8} \right) \prod_{i=1}^n
 \mathbb{E}_{ h \sim  \mu_{i} } 
 \left\{ \exp\left[ -  \zeta  \ell(h)  \right] \right\}    
   \text{, using~\eqref{hoeffding}}
 \\
 & =  
  \exp\left(- \frac{n C^2 \zeta^2}{8} \right) \prod_{i=1}^n
   \int 
 \left\{ \exp\left[ -  \zeta  \ell(h)  \right] \right\}    \mu_{i} ({\rm d} h)
  \\
 & = 
   \exp\left(- \frac{n C^2 \zeta^2}{8} \right) \prod_{i=1}^n
 \int 
 \frac{\exp\left\{ -
   \zeta \sum_{u=1}^i  \ell(h_u) 
   \right\}
   }{W_{i}} \mu_1({\rm d} h)
    \text{, using~\eqref{renorm}}
 \\
 & = 
   \exp\left(- \frac{n C^2 \zeta^2}{8} \right) \prod_{i=1}^n
   \frac{W_{i+1}}{W_{i}}
 = \exp\left\{ \frac{n C^2 \zeta^2}{8}
 \right\} W_{n+1}.
\end{align*}
So
\begin{align*}
 \sum_{i=1}^n \mathbb{E}_{h \sim  \mu_{i} }[ \ell(h) ]
 & \leq 
  - \frac{\log W_{n+1}}{ \zeta}
     + \frac{n C^2 \zeta}{8}
 \\
 & = -
  \frac{\log \int \exp\left[-\zeta  \sum_{i=1}^n  \ell(h_i)  \right] \mu_1 ({\rm d}h)}{ \zeta}
     + \frac{n C^2 \zeta}{8}
\end{align*}
and finally we use \cite[Equation (5.2.1)]{catoni2004statistical}
which states that
$$
 - \frac{\log \int \exp\left[-\zeta \sum_{i=1}^n  \ell(h_i)  \right] \mu_1 ({\rm d}h)}{ \zeta}
 = \inf_{\nu} \left\{ \mathbb{E}_{ h_i \sim \nu}\left[  \sum_{i=1}^n  \ell(h_i)     \right] 
 + 
 \frac{\mathcal{K}(\nu,\mu_1 )}{\zeta} \right\}.
$$ 

Therefore, for each $t$ and given a $\theta$, we obtain a general bound
\begin{align}
\label{cor_generalbound}
 \frac{1}{n} \sum_{i=1}^{n} \hat{\ell}_{t,i}
 \leq
 \inf_{\nu} 
       \mathbb{E}_{h_t\sim \nu}\Bigg\{
       \frac{1}{n} \sum_{i=1}^{n}
        \ell ( h_t(\theta^\top x_{t,i}),y_{t,i})
     + \frac{\zeta C^2}{8} 
     + \frac{\mathcal{K}(\nu,\mu_1)}{\zeta n} \Bigg\}.
 \end{align}

Put
\begin{align*}
h_t^* &:= \arg \inf_{h_t \in\mathcal{H}_{S,C_2+1} }
\frac{1}{n} \sum_{i=1}^{n}
 \ell  ( h_t(\theta^\top x_{t,i}),y_{t,i}) .
\end{align*}
We define 
\begin{align*}
\| h \|_S =  \sum_{j=1}^Sj|\beta_j|  , \forall h \in \mathcal{H}_{S,C_2+1} .
\end{align*}
and let
\begin{align*}
\nu_{\gamma} = \mathbf{1} (\|h- h^*_t \|_S \leq \gamma) \mu_1 (dh).
\end{align*}
The bound in \eqref{cor_generalbound} becomes
\begin{align*}
 \frac{1}{n} \sum_{i=1}^{n} \hat{\ell}_{t,i}
 \leq
 \inf_{\gamma} 
       \mathbb{E}_{h_t\sim \nu_{\gamma}}\Bigg\{
       \frac{1}{n} \sum_{i=1}^{n}
        \ell ( h_t(\theta^\top x_{t,i}),y_{t,i})
     + \frac{\zeta C^2}{8} 
     + \frac{\mathcal{K}(\nu_{\gamma},\mu_1)}{\zeta n} \Bigg\}.
 \end{align*}

Under the condition on the loss, we have
\begin{align*}
\Big| \ell ( h^*_t(\theta^\top x_{t,i}),y_{t,i})
 -
 \ell ( h_t(\theta {}^\top x_{t,i}),y_{t,i}) \Big| 
 &      \leq
L_1\,  \Big|  h^*_t(\theta^\top x_{t,i})
 -  h_t(\theta {}^\top x_{t,i}) \Big|
 \\
& \leq
L_1 \sup_{z}  | h^*_t(z) -  h_t(z) |
\\
& \leq
L_1 \gamma.
\end{align*}
Using the Lemma 10 in \cite{alquier2013sparse}, we have 
$$\mathcal{K}(\nu_{\gamma},\mu_1) \leq S\log\frac{(C_2 +1)}{\gamma} .$$
\\
Thus we obtain
\begin{align*}
 \frac{1}{n} \sum_{i=1}^{n} \hat{\ell}_{t,i}
 -   \inf_{h_t \in\mathcal{H}_{S,C_2+1} }\frac{1}{n}\sum_{i=1}^{n}
 \ell(h_t(\theta^\top x_{t,i}),y_{t,i})
 \leq
 \inf_{\gamma} 
\Bigg\{    L_1 \gamma  + \frac{\zeta C^2}{8} 
     + \frac{S\log\frac{(C_2 +1)}{\gamma} }{\zeta n} \Bigg\}.
\end{align*}

By choosing $\gamma = 1/\sqrt{n}$ and then the optimum is reached at $\zeta = \sqrt{\frac{ 8S}{C^2 n}}$
\begin{align*}
 \frac{1}{n} \sum_{i=1}^{n} \hat{\ell}_{t,i}
 -   \inf_{h_t \in\mathcal{H}_{S,C_2+1} }\frac{1}{n} \sum_{i=1}^{n}
 \ell(h_t(\theta^\top x_{t,i}),y_{t,i})
 \leq
\frac{ L_1 }{\sqrt{n}} + \frac{C\sqrt{S}}{2\sqrt{2n}} 
     + \frac{C\sqrt{S} \log[(C_2 +1)\sqrt{n}]}{2\sqrt{2n}}  .
\end{align*}
This completes the proof.
\end{proof}

\section*{Acknowledgments}

TTM is supported by the Norwegian Research Council grant number 309960 through the Centre for Geophysical Forecasting at NTNU.

%
%
\bibliographystyle{splncs03}

\end{document}